\documentclass[runningheads]{llncs}
\usepackage{graphicx}
\usepackage{amsmath}
\usepackage{cite}  % allows for indexgeneration
\usepackage{graphicx}
\usepackage{makecell}
\usepackage{amsfonts}
\usepackage[caption=false]{subfig}
\usepackage{multirow}
\usepackage{amssymb}
\usepackage{color}
\begin{document}
\title{OASIS: One-pass aligned Atlas Set for Image Segmentation}

\author{Qikui Zhu\textsuperscript{1}, Bo Du\textsuperscript{1}, Pingkun Yan\textsuperscript{2}}


%
\authorrunning{Zhu, Du, Ping}

\institute{1 School of Computer Science, Wuhan University, Wuhan, China \\ 2 Department of Biomedical Engineering and the Center for Biotechnology and Interdisciplinary Studies at Rensselaer Polytechnic Institute (RPI), Troy, NY, USA}
\maketitle              % typeset the header of the contribution
\begin{abstract}
Medical image segmentation is a fundamental task in medical image analysis. Despite that deep convolutional neural networks have gained stellar performance in this challenging task, they typically rely on large labeled datasets, which have limited their extension to customized applications. By revisiting the superiority of atlas based segmentation methods, we present a new framework of One-pass aligned Atlas Set for Images Segmentation (OASIS).
To address the problem of time-consuming iterative image registration used for atlas warping, the proposed method takes advantage of the power of deep learning to achieve one-pass image registration. In addition, by applying label constraint, OASIS also makes the registration process to be focused on the regions to be segmented for improving the performance of segmentation. Furthermore, instead of using image based similarity for label fusion, which can be distracted by the large background areas, we propose a novel strategy to compute the label similarity based weights for label fusion. Our experimental results on the challenging task of prostate MR image segmentation demonstrate that OASIS is able to significantly increase the segmentation performance compared to other state-of-the-art methods.

%by benefiting from the spatial relationships between anatomical structures and the similar context with atlas.
%\keywords{First keyword  \and Second keyword \and Another keyword.}
\end{abstract}
\section{Introduction}
Accurate segmentation of medical images is of great significance for clinical practice, especially for disease diagnosis and treatment planning. For instance, prostate image segmentation is useful for prostate cancer radiotherapy planning and guidance \cite{shen2003segmentation}.
%Segmentation of the liver and tumors plays an important role in hepatocellular carcinoma diagnosis\cite{heimann2009comparison}. %
In the past several years, deep convolutional neural networks (CNNs) have obtained impressive progress in medical image segmentation due to their powerful hierarchical representation ability. However, training such networks usually requires large amount of training data with corresponding segmentation label, which is difficult to obtain due to the needed expertise and highly intensive labor.
%\cite{Automaticmeyer,ghafoorian2017transfer,mortazi2017cardiacnet}
Given relative small datasets, degraded performances are usually observed when segmenting anatomical structures with large appearance and shape variation.

%To deal with the challenge of lacking training data, data augmentation and transfer learning\cite{ShinTMI} are two commonly adopted strategies. Data augmentation employs rotation, scaling or flipping operation to artificially increase the number of training samples. On the other hand, transfer leaning tries to utilize other related datasets to help train the networks. \blue{ In fact, both of them aim to exploit more useful information for network training and further improve the ability of generalization and robustness. However, the strategy of data augmentation are more tend to increase the network's ability of rotation and translation invariant, and transfer learning is more meaningful in weight initialization and always meet the challenge of domain shift between different dataset. Furthermore, both of them also increase the load of computing and the complexity of network.}
%\cite{shan2018correction,van2015transfer,hoo2016deep}

Before the deep learning era, atlas based segmentation methods were often used for medical image segmentation \cite{atlas_theory_tmi19, andreasen1996automatic,yang2016deep}. That is based on the fact that structures and organs share large similarity in appearance and shapes across subjects.
%Based on this characteristic, multi-atlas segmentation methods\cite{yang2016deep,atlas_theory_tmi19} be developed,
Atlas based methods segment images by fusing the label of similar images, i.e. atlases, through image alignment \cite{yan2015label}. Such segmentation algorithms have two major steps, atlas selection and label fusion. The former is to select a few most similar images from the training data for a target image as atlases, which relies on similarity measurement and ranking. Label fusion is to fuse the warped atlas labels after image registration to segment the target image.
Several key factors have limited the effectiveness of atlas based segmentation methods. Fist, each atlas has to be registered with the target image. Classical iterative deformable registration can be computational intensive, which slows down the entire segmentation process. Second, the registration considers the whole image equally, which results in the registration performance being dragged by less relevant background. Third, label fusion is based on the similarity between the target and registered images rather than the region of interest, which further degrades the segmentation accuracy.

In this paper, to infuse the power of deep learning into atlas based methods for improved performance, we propose OASIS -- One-pass aligned Atlas Set for Image Segmentation.
%\textbf{(iii)}
%Typically each atlas has to be registered with each target image every time, which is time-consuming and computational intensive.
To overcome the problem of time-consuming iterative image registration, the proposed OASIS employs a label constrained spatial transform model (STM) for one-pass image registration, which also allows the alignment process to be focused on the regions to be segmented and further improve the performance.
%\item
%\textbf{(ii)}
Instead of fusing atlas labels weighted by using similarity between atlas images and test image, to reduce the distraction of background on label fusion, we propose a novel fusion strategy taking the contribution of each atlas label. It is achieved by measuring the similarity of registered labels, which makes the label fusion focusing on the regions to be segmented rather than the whole image. The experimental results show that the proposed fusion strategy significantly enhances the accuracy of segmentation.
%\item
It is worth noting that OASIS is a general framework and can be easily extended to other medical image analysis tasks, especially those with limited training data.
%\end{enumerate}
%\blue{May be we can remove this? The rest of this paper is organized as follows.  Section~\ref{sec:Method} describes the proposed OASIS in detail. Section~\ref{sec:Experiments} presents and discusses the performance of the proposed method through various experiments on prostate MRI image segmentation. Finally, several concluding remarks are drawn in Section~\ref{sec:conclusions}.}

\begin{figure}[t]
\center{\includegraphics[width = \textwidth] {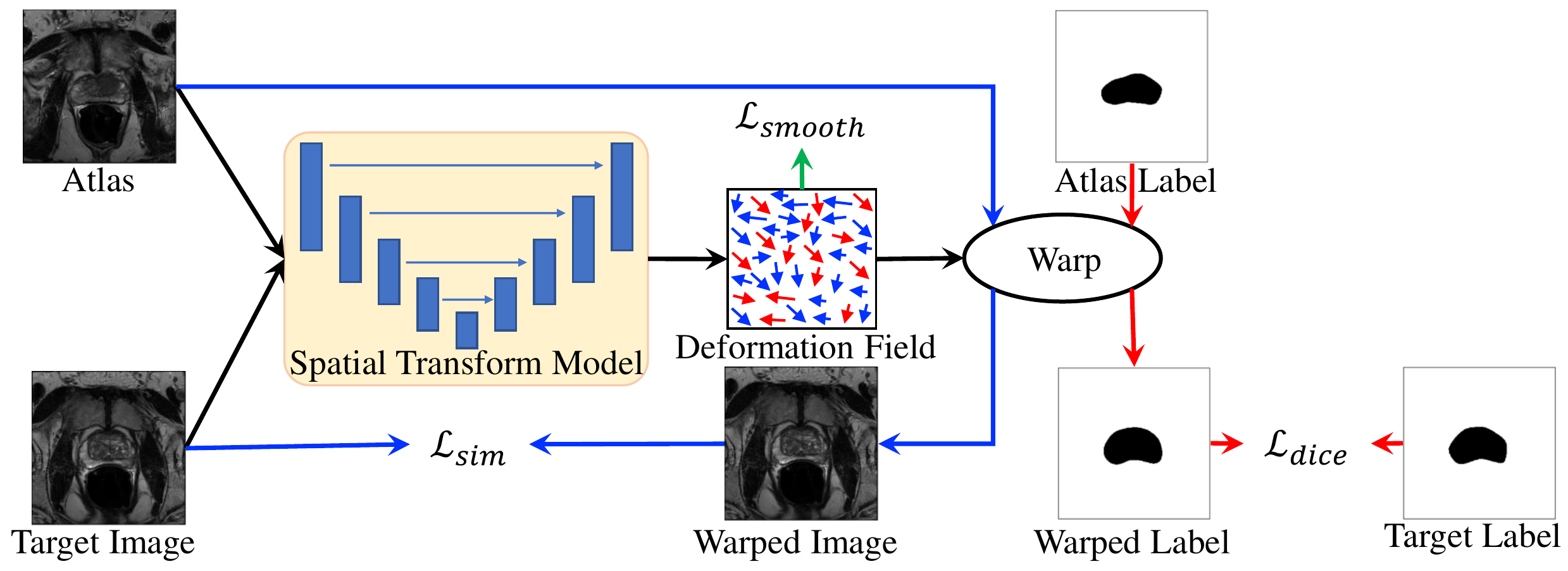}}
\caption{Label constrained spatial transform model for one-pass atlas alignment.}
\label{fig:ProposedModel}
\end{figure}

\section{Aligning and Fusing Atlases for Image Segmentation}
\label{sec:Method}

The proposed OASIS framework consists of two major steps -- spatial transform based atlas alignment and weighted label fusion for segmentation. The former aims to learn the spatial transforms between images for atlas warping. The latter computes the weight of atlases label and then fuses them to get the final segmentation. Details of these two steps are presented in the following subsections.

\subsection{Spatial Transform based Atlas Alignment}

In our work, we leverage the recent progress in deep learning based image registration, especially VoxelMorph \cite{VoxelMorph2019,balakrishnan2018unsupervised}, which registers a pair of images by jointly optimizing an image similarity loss and a displacement field smoothness term, as spatial transform model (STM) for atlas warping. To make the spatial transform model (STM) focus on the region to be segmented for improving registration accuracy, we constrain the STMs with labels of atlas and target image by supervising the training process of learning as shown in Fig.~\ref{fig:ProposedModel}. Formally, let ${\{\mathbf{x}_a,\mathbf{y}_a\}} = \{ (x_a^i,y_a^i)\left| {i = 1,...,n} \right.\}$ represent a set of atlases and the corresponding labels, ${\{x_t,y_t\}}$ be a target image with corresponding label, and $\phi$ denote the deformation filed learned by STM. The total loss of label constrained STM for each atlas-target pair can be written as
\begin{equation}\label{Dloss}
 \mathcal{L}(x_a^i, y_a^i, x_t, y_t, \phi ) =   \alpha \mathcal{L}_{sim}(\phi(x_a^i), x_t) + \beta \mathcal{L}_{dice}(\phi(y_a^i), y_t) + \gamma \mathcal{L}_{smooth}(\phi),
\end{equation}
where $\alpha$, $\beta$, and $\gamma$ are positive weighting parameters. $\phi(x_a^i)$ and $\phi(y_a^i)$ are the warped atlas and label by the deformation filed $\phi$.

The loss term $\mathcal{L}_{sim}(\phi(x_a^i), x_t)$ measures the similarity between warped atlas image $\phi(x_a)$ and target image $x_t$. In our work, it is defined by using cross correlation as
\begin{equation}
\mathcal{L}_{sim}  =  - \frac{\sum_{i=1}^{N}\left [ \left ( \phi(x_a^i)-\overline{\phi(x_a^i)} \right )\left ( x_t-\overline{x_t} \right ) \right ]}{\sqrt{\sum_{i=1}^{N} \left ( \phi(x_a^i)-\overline{\phi(x_a^i)} \right )^{2}}\sqrt{\sum_{i=1}^{N}\left ( x_t-\overline{x_t} \right )^{2}}},
\label{NCC}
\end{equation}
where $N$ denotes the number of pixels in an image.
The loss term $\mathcal{L}_{dice}(\phi(y_a), y_t)$ evaluates the similarity between warped atlas label $\phi(y_a)$ and target label $y_t$, where Dice score is employed for this purpose
\begin{equation}
\mathcal{L}_{dice}  = - 2 * \frac{{\phi (y_a^i) \cap (y_t)}}{{\left| {\phi (y_a^i)} \right| + \left| { (y_t)} \right|}}.
\label{Dice}
\end{equation}
Similarly to VoxelMorph \cite{VoxelMorph2019}, $\mathcal{L}_{smooth}(\phi)$ is employed to regularize $\phi$ for obtaining spatially smoothing deformation as
\begin{align}
\mathcal{L}_{smooth}  =\sum\limits_{i \in \Omega } {{{\left\| {\nabla \phi } \right\|}^2}}.
\label{smooth}
\end{align}
%The image similarity can be computed by  many formulations, such as cross correlation (CC), normalized cross correlation (NCC) or mean squared error (MSE). The smoothness regularization which often based on gradients information. In our work, we employ cross correlation as image similarity loss $\mathcal{L}_{sim}$, dice score is employed as $\mathcal{L}_{dice}$ for measuring the accuracy of warped atlas label and target label and using a diffusion regularizer on $\phi$ spatial gradients as smooth loss $\mathcal{L}_{smooth}$. The loss terms $\mathcal{L}_{sim}$, $\mathcal{L}_{dice}$ and $\mathcal{L}_{smooth}$ are defined as follows

\subsection{Weighted Label Fusion}

\begin{figure}[t]
	\center{\includegraphics[width =.9\textwidth] {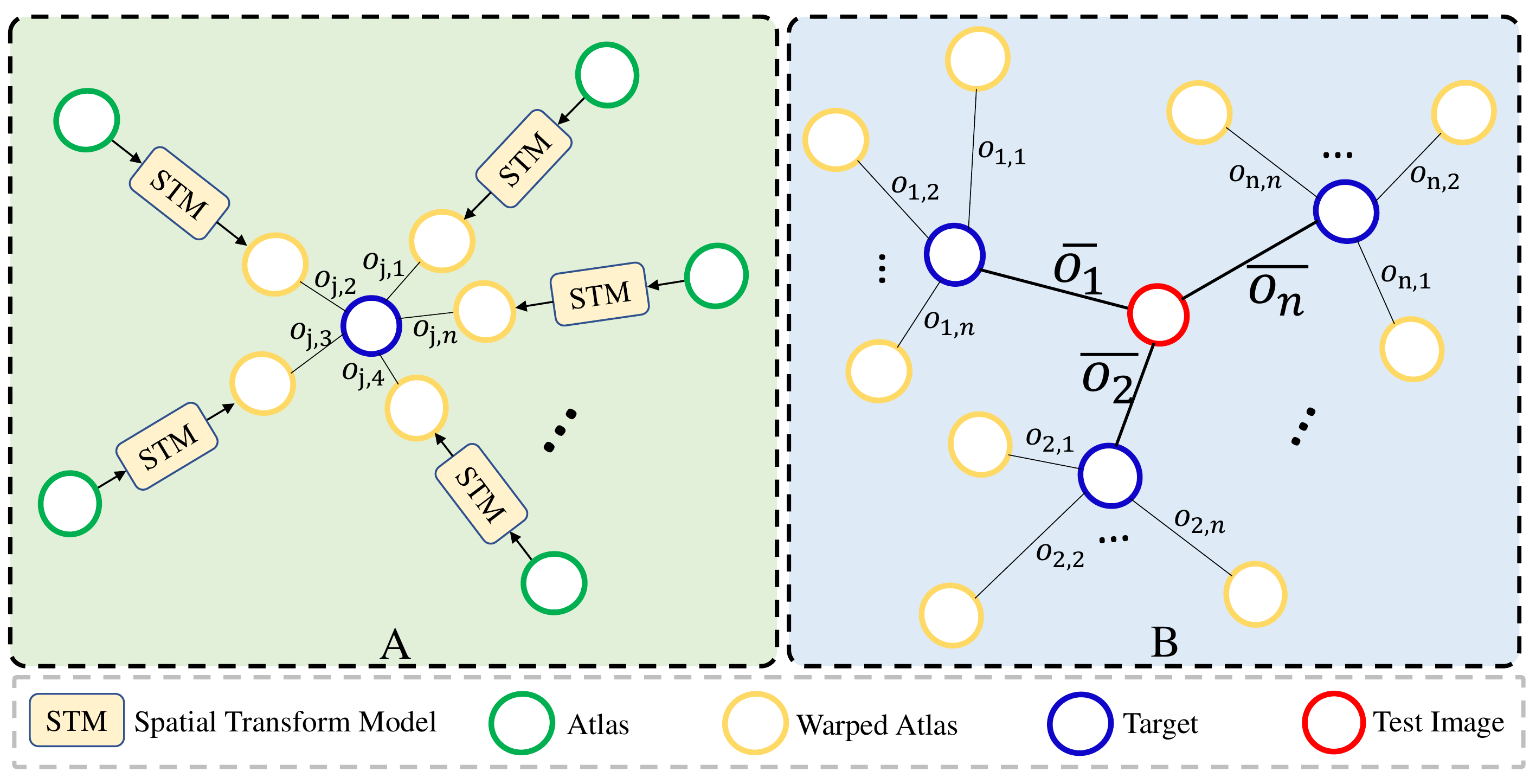}}
	\caption{Overview of the process for label fusion weight computation.}
	\label{fig:Fusing}
\end{figure}

Label fusion is a key step in predicting the test image label. Typically, the weights of atlas labels are computed based on the similarity between the atlas and test images. However, a major weakness is that the weight is determined by both the region to be segmented and also the large background areas, where the weights do not accurately reflect the similarity between the test label and atlas labels. To overcome this problem, in this paper, we propose a novel weight-computing strategy, which utilizes the label overlapping accuracy (LOA) defined as $o = \{ o_{j,k}\left| {j,k = 1,...,n} \right.\}$. It evaluates the contribution of an atals to the target image and computed by the similarity between each warped atlas label $k$ and the target label $j$ as shown in Fig.\ref{fig:Fusing}(A).

Besides the similarity between target label and warped atlas label, the value of $o_{j,k}$ also represents the probability of each pixel belongs to region to be segmented. Furthermore, the average accuracy $\bar o_j = \frac{1}{n}\sum\limits_{j = 1}^n {{o_{j,k}}}$ is the expected value of a target label coming from each atlas label. Relying on this evaluation, during test, the weight of target atlas labels for test image can be obtained. The process is shown in Fig.~\ref{fig:Fusing}(B). In our work, we measure LOA by Dice score, which quantifies the normalized overlap between two labels as given by Eqn.~(\ref{Dice}).

%Label fusion is a key step for predicting the target label, in this paper, we proposed a novel fusion strategy relying on neighboring-evaluation.
%The atlas can be considered as a neighborhood for target image, each neighborhood can score the target image. And the average score from all of neighborhoods can represent the contribution of target image for its neighborhoods. Fig.\ref{fig:Fusing}(A) presents the process of neighboring evaluation.
%When each atlas label with corresponding target label through the trained spatial transform model, we can obtain transformed atlas label. By computing the accuracy $ \{ w_{i,j}\left| {i,j = 1,...,n} \right.\}$
%between target label with transformed atlas label, the contribution of target image for each transformed atlas can be evaluated. Correspondingly, the average accuracy $\bar w_i = \frac{1}{n}\sum\limits_{j = 1}^n {{w_{i,j}}}$ from all of atlas label represents the contribution of target image for each atlas. Relying on this evaluation, during test, the contribution of target image for test image can be evaluated by the atlas of target image  as shown in  Fig.\ref{fig:Fusing}(B). In our work, we measure the accuracy using the dice score, which quantifies the label overlap between two labels, the formula of dice as shown in Eqns.~(\ref{Dice}).
 %\begin{align}\label{weight}
%\bar w_i = \frac{1}{n}\sum\limits_{j = 1}^n {{w_{i,j}}}
%\end{align}

\section{Experiments}
\label{sec:Experiments}
\subsection{Experimental Data and Preprocessing}

In our work, the MICCAI 2012 Prostate MR Image Segmentation (PROMISE12) challenge\footnote{https://promise12.grand-challenge.org/} dataset, a benchmark for evaluating algorithms of segmenting the prostate from MR images, is used for the evaluation. This dataset includes in total 50 transversal T2-weighted MR images of the prostate, which are a representative set of the types of prostate MR images from multiple vendors and have different acquisition protocols and variations in voxel size, dynamic range, position, field of view and anatomic appearance. The corresponding ground truth segmentation were annotated and checked by radiological residents.

In our experiments, due to the large variation of voxel size, resolution, dynamic range, position, and field of view in the PROMISE12 dataset, we first resized all the images into a fixed size of 320 $\times$ 320 pixels. We then normalized each dataset to have zero mean and unit variance. It is worth noting that we do not employ data augmentation in our experiment, since we would like to prove that the information from atlas is sufficient and effective for the segmentation.  Although we have resized all images into a same size, the feature distributions between different subjects are still various. Therefore, in our work, we employ pre-trained DenseNet-121~\cite{huang2017densely} as a feature extractor, which maps each image to a feature vector with length of 1000. Then the normalized Euclidean distance is employed as similarity measure to evaluate the distance between images for atlas picking. In our experiment,  we select $n = 6$ atlases for each image and set the weighted coefficients $\alpha = 1.0$, $\beta = 1.0$ and $\gamma = 0.01$ in Eqn.~(\ref{Dloss}).

The proposed method is implemented using the open source deep learning library Keras \cite{keras}. The label constrained STM  is trained end-to-end with Adam optimization method and the learning rate is set to $1.0 \times 10^{-5}$. The experiments were carried out on a NVIDIA GTX 1080ti GPU with 11GB memory, we chose  16 as batch size.

\subsection{Experimental Results}

The ground truth of testing data is held out by organizers and the challenge organizers, who require that each submission is accompanied by a publication on arXiv or any other official preprint server. The requirement is not in accordance with the rules of MICCAI\footnote{http://www.miccai2019.org/information/information-authors/} anonymous submission. Thus we conducted out experiments via a standard 5-fold cross validation scheme in this paper. %under 50 transversal T2-weighted MR images.
The evaluation metrics used in our experiment is Dice Similarity Coefficient (DSC), the absolute relative volume difference (aRVD) and Hausdorff Distance (HD). All the evaluation metrics are calculated in 3D. %which is calculated in 3D.average over the shortest distance between the boundary points of the volumes (ABD)

\begin{figure}[t]
\center{\includegraphics[width = 0.7\textwidth] {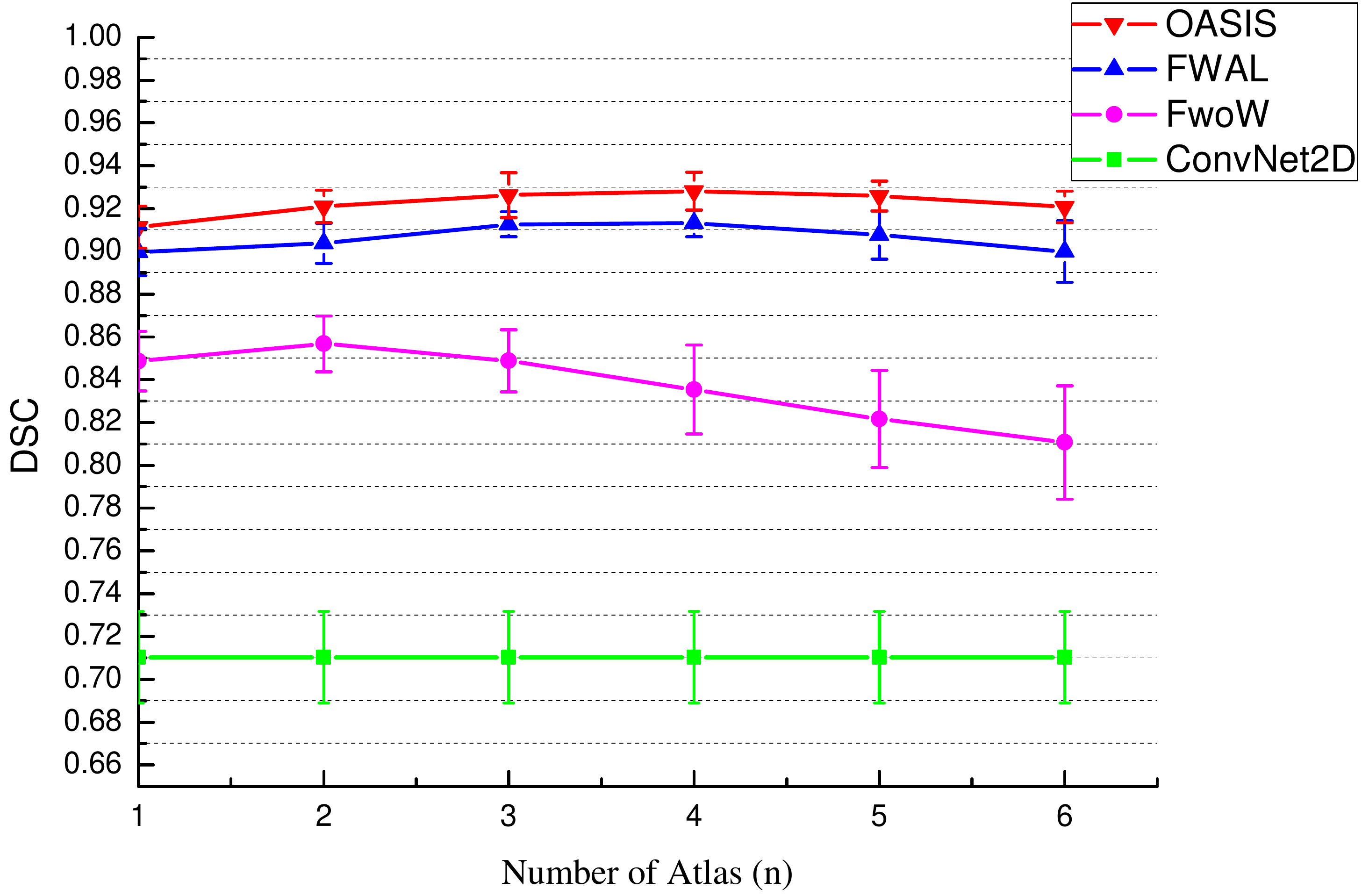}}
\caption{Segmentation performance of different strategies under various number of atlas.}
\label{fig:ComRe}
\end{figure}

\begin{figure}[t]
\center{\includegraphics[width=\textwidth] {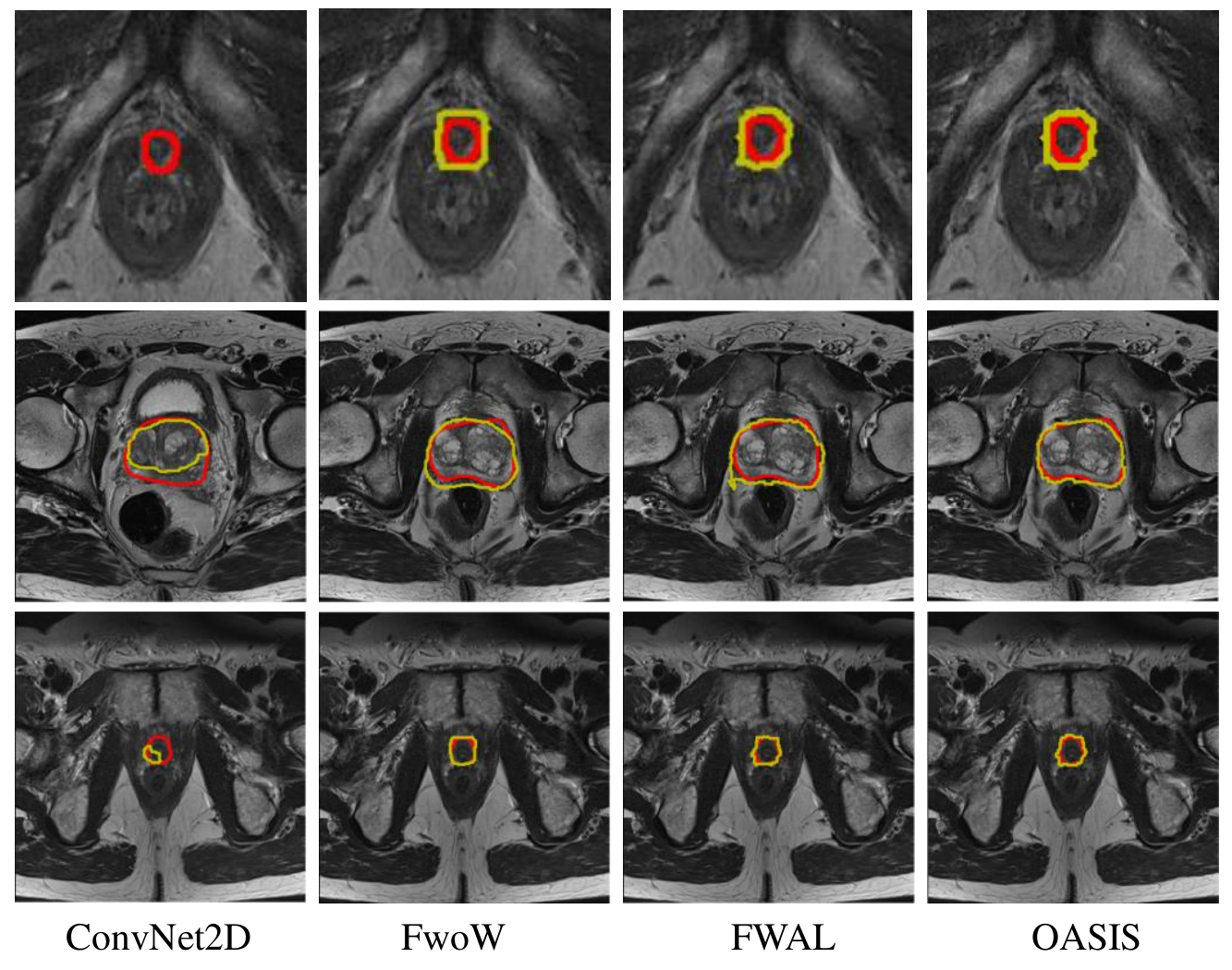}}
\caption{Sample segmentation results of different methods. The red and yellow contours indicate ground truth and segmentation results, respectively.}
\label{fig:SegRe}
\end{figure}

In our experiments, to demonstrate the effectiveness of label constrained STM and the proposed label fusion strategy, we analyze the influence of each part on the segmentation results through ablation studies. Three other baseline methods are included as follows.
\emph{(1) Fusing without Warping (FwoW)}: The segmentation is achieved by directly fusing the atlas labels without warping. The weight of each atlas label is computed by the similarity between the atlas and test images.
\emph{(2) Fusing Warped Atlas Label (FWAL)}: Different from (1), here the segmentation is acquired by fusing the warped atlas labels and the weights of the warped atlas labels are computed based on the similarity between the warped atlas and test images.
\emph{(3) ConvNet2D}: We also include the convolutional neural network based model for comparison, which is an implementation of ConvNet~\cite{yu2017volumetric}. The method once ranked the first for this challenge in 2016. Since our proposed framework is 2D based, we replace the 3D convolutional layers inside ConvNet with 2D convolutions.
%\emph{(4) OASIS}: The proposed segmentation framework.

In addition to evaluating these models over the entire prostate as shown in Fig.~\ref{fig:ComRe}, same as the challenge, we also calculated the performance of each model in the apex and base areas of the prostate.  The apex and base are determined by dividing the prostate into three approximately equally sized parts along the axial direction (the first 1/3 as apex and the last 1/3 as base). We also employ 3D ConvNet~\cite{yu2017volumetric} for comparing.  %{Note that the segmentation results of ConvNet were obtained directly from the challenge website\footnote{https://promise12.grand-challenge.org/evaluation/results/} and the segmentation result of each atlas based model comes from the best condition as shown in Fig.\ref{fig:ComRe}.}
The results of each method are shown in Table~\ref{tab:prostate_comparison}. As it can be seen, the segmentation results of our OASIS were the best not only for whole prostate segmentation, but also in the base and apex areas, which further demonstrates the effectiveness of the proposed approach. Actually Table~\ref{tab:prostate_comparison} shows that OASIS has more performance gain in the base and apex (4.1\% in Dice) than the overall whole prostate segmentation (3.4\% in Dice).

\begin{table}[t]
  \centering
%  \fontsize{8}{11}\selectfont
  \caption{Quantitative evaluation results of proposed models and other methods on PROMISE12 challenge dataset.}
  \label{tab:prostate_comparison}
    \begin{tabular}{l|ccc|ccc|ccc}
    \hline
    %\Xhline{1.2pt}
    \multirow{2}{*}{Method}&
    \multicolumn{3}{c|}{aRVD [\%]} & \multicolumn{3}{c|}{HD [mm]} & \multicolumn{3}{c}{DSC [\%]}\cr
    \cline{2-10} & \ Apex& \ Base& \  Whole&  \ Apex&  \ Base&  \ Whole&  \ Apex&  \ Base&  \ Whole\cr
	\Xhline{1.2pt}

    ConvNet2D & 38.21 & 37.91 & 32.29 & 9.45 & 14.88 & 24.97  & 67.93 & 63.73 &  71.03   \cr

    FwoW-2 & 17.43 & 15.64 & 16.26 & 6.32 & 6.64 & 14.16 & 84.50 & 86.53 & 85.68 \cr

    FWAL-4 & 9.99 & 9.98 & 13.14 & 5.35  & 6.11 & 10.51 & 90.82 & 90.83 & 91.31  \cr

    ConvNet3D\cite{yu2017volumetric} & 15.18 & 11.04 & 6.95 & \textbf{4.17} & 5.22 & \textbf{5.13} & 86.81 & 86.42 & 89.43 \cr

    OASIS-4  & \textbf{9.82} & \textbf{9.92} & \textbf{6.63} & 4.61 & \textbf{5.05} & 6.59 & \textbf{90.90} & \textbf{90.87} & \textbf{92.81} \cr
  \hline
  %		\Xhline{1.2pt}

    \end{tabular}
\end{table}

Fig.~\ref{fig:ComRe} shows the segmentation performance  on the over the entire prostate of the methods described above. It can be seen that ConvNet2D has a poor performance, which is even worse than the segmentation results obtained by FwoW. The major reason is that the limited training data makes it difficult to get the network fully trained for segmentation.
Compared with FwoW, FWAL achieved better accuracy, which demonstrates the advantage of our proposed label constrained STM in image registration.
Remarkably, our proposed OASIS achieved the best performance, which proves the effectiveness of our proposed method including the weight-computing strategy. In addition, compared with FWAL and FwoW, the performance of OASIS decreases much less when the number of atlases increase. That may be because the proposed weight-computing strategy estimates better weights for fusing the labels.
Some qualitative segmentation results are shown in Fig.~\ref{fig:SegRe}. It is observed that OASIS can produce more accurate segmentation results than other methods. In the same time, ConvNet2D has poor performance in the base and apex areas, where the atlas based methods work better. The results also indicate that while CNN based models have achieved remarkable success in many medical image segmentation applications, their performance may be sub-par when dealing with challenging tasks and the available training data are limited.

%Fig.\ref{fig:ComRe} also shows the segmentation performances of atlas based methods under various number of atlas. As it can be seen that when the employed atlas number is \textcolor{red}{3}, the proposed framework obtains the highest DSC accuracy \textcolor{red}{87.6\%}. After that the accuracy declines slowly. Remarkably, the lowest accuracy obtained by atlas is higher than the result from segmentation network, which demonstrates the advantage of atlas be employed.

%\begin{table}
%	\caption{Quantitative evaluation results of proposed framework and other methods on PROMISE12 challenge dataset.}
%	\centering
%	 \label{tab:prostate_comparison1}
	%	\fontsize{8.5}{10}\selectfont
%	\begin{tabular}{c|c|c|c|c}
%		\hline
%		\centering
%		Method &  Data Augmentation & HD [mm] & aRVD [\%] & DSC [\%]\\
		%\hline
%		\Xhline{1.2pt}
%		SegNet  & $\times$   & 17.25 & 0.18 & 86.89\\
%		\hline	
%		SA - 2  & $\times$   & 17.25 & 0.18 & 86.89\\
%		\hline			
%		ConvNet\cite{yu2017volumetric} & \checkmark & - & -  & 86.93 \\
%		\hline
%		TA - 2  & $\times$   & 10.35  & 0.14  & 89.74\\
%		\hline	
%		OASIS - 4 & $\times$   & \textbf{8.58}  & \textbf{0.06} & \textbf{92.24}\\
%		\hline
	%	\Xhline{1.2pt}
%	\end{tabular}
%	\label{Tabel_Loss}
	
%\end{table}

\section{Conclusion}
\label{sec:conclusions}
In this paper, inspired by atlas based segmentation methods, an effective medical image segmentation framework: One-pass aligned Atlas Set (OASIS) is proposed for medical image segmentation. Benefits from the superiority of deep learning, a one-pass image registration model is designed to overcome the problem of time-consuming iterative image registration in atlas based methods. Furthermore, instead of determining the weight of atlas labels by using similarity between atlas images and test image, we proposed a novel  fusion  strategy relying on the contribution of each atlas label, which can further reduce the  distraction of  background  on  label  fusion and make the segmentation results rely on the region to be segmentation rather than the whole image. Extensive experiments on an open challenge dataset (PROMISE12) demonstrate that our proposed OASIS can achieve superior results compared with other state-of-the-art methods under limited training data. In our future work, we will extend the proposed method to other segmentation tasks on different organs and modalities.
% ------------- %
\bibliographystyle{splncs04}
\bibliography{references}

\end{document}